\documentclass[10pt,twocolumn,letterpaper]{article}

\usepackage{cvpr}              % To produce the CAMERA-READY version
\newcommand{\xvec}{\vec{x}}
\newcommand{\omegavec}{\vec{\omega}}
\newcommand{\uvec}{\vec{u}}
\newcommand{\svec}{\vec{s}}

\definecolor{cvprblue}{rgb}{0.21,0.49,0.74}
\definecolor{warning}{RGB}{255, 69, 0}
\usepackage[pagebackref,breaklinks,colorlinks,allcolors=cvprblue]{hyperref}
\usepackage{layouts}

\usepackage{colortbl}
\usepackage{xcolor}
\usepackage{siunitx}
\usepackage{float}

\def\paperID{4097} % *** Enter the Paper ID here
\def\confName{CVPR}
\def\confYear{2025}

\title{Event fields: Capturing light fields at high speed, resolution, and dynamic range}

\author{Ziyuan Qu\\
Dartmouth College
\and
Zihao Zou\\
University of North Carolina
\and
Vivek Boominathan\\
Rice University
\and
Praneeth Chakravarthula\\
University of North Carolina
\and
Adithya Pediredla\\
Dartmouth College
}

\begin{document}
\twocolumn[{
\renewcommand\twocolumn[1][]{#1}%
\maketitle
\begin{center}
    \vspace{-15pt}
    \centering
        \includegraphics[width=\linewidth]{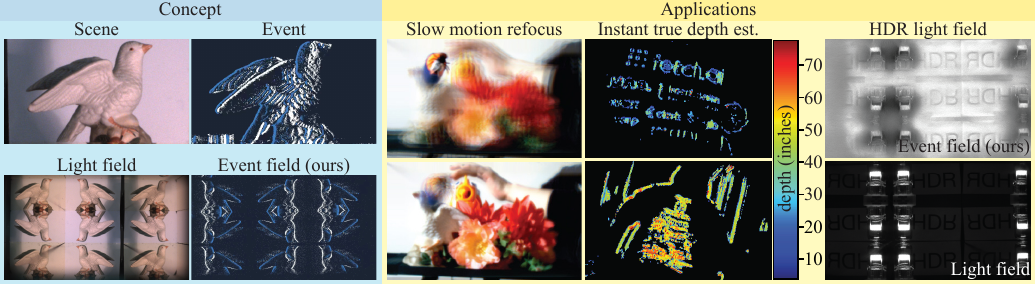}
        % \vspace{-2em}
    \captionof{figure}
    {We introduce \emph{Event Fields}, a novel framework that leverages event cameras to capture light fields at high speed, resolution, and dynamic range using innovative optical designs. 
    Event fields integrate the unique advantages of event sensing---such as low latency, data efficiency, and power-saving capabilities---with the detailed angular information provided by light field sensing.
    % We propose two designs to either spatially (using a Kaleidoscope) or temporally (using a Galvanometer) multiplex angular information in the captured events. 
    % To enable efficient event field capture, we propose two designs, which spatially and temporally multiplex light-field information in the captured events. 
    % We built a kaleidoscopic and galvanometer design
    We demonstrate the versatility of the event fields through three application scenarios: slow motion refocus (SloMoRF), instant true depth estimation, and high dynamic range (HDR) light field imaging.}
    % {We introduce \emph{Event Fields}, a novel framework that leverages event cameras to capture high-speed light fields using innovative optical designs. Event fields combine the unique advantages of event sensing, including sparseness, high speed, and high resolution, with the rich angular and color information of light fields. To enable efficient event field capture, we propose two hardware designs: the kaleidoscope and the galvanometer. We demonstrate the versatility of the event field through three application scenarios: slow motion refocus (SloMoRF), instant true depth estimation, and high dynamic range (HDR) light fields.}
    \label{fig:sec01_01_teaser}
    \vspace{5pt}
\end{center}
}]

% \begin{center}
%   \centering
%   \includegraphics[width=\linewidth]{fig/pdf_file/sec01_01_teaser.pdf}
%   \caption{\textbf{(Teaser)}}
%   \label{fig:sec01_01_teaser}
% \end{center}
\begin{abstract}
Event cameras, which feature pixels that independently respond to changes in brightness, are becoming increasingly popular in high-speed applications due to their lower latency, reduced bandwidth requirements, and enhanced dynamic range compared to traditional frame-based cameras.
Numerous imaging and vision techniques have leveraged event cameras for high-speed scene understanding by capturing high-framerate, high-dynamic range videos, primarily utilizing the temporal advantages inherent to event cameras. 
Additionally, imaging and vision techniques have utilized the light field---a complementary dimension to temporal information---for enhanced scene understanding. 

In this work, we propose ``Event Fields", a new approach that utilizes innovative optical designs for event cameras to capture light fields at high speed. 
We develop the underlying mathematical framework for Event Fields and introduce two foundational frameworks to capture them practically: 
% one based on temporal derivatives and the other on angular derivatives.
% one leveraging temporal derivatives and another using angular derivatives. 
spatial multiplexing to capture temporal derivatives and temporal multiplexing to capture angular derivatives.
To realize these, we design two complementary optical setups---one using a kaleidoscope for spatial multiplexing and another using a galvanometer for temporal multiplexing. 
We evaluate the performance of both designs using a custom-built simulator and real hardware prototypes, showcasing their distinct benefits. 
% We demonstrate new high-speed capabilities made possible by Event Fields, such as post-capture refocusing and accurate depth capture of high-speed scenes, and dynamic focus tracking across diverse scenarios.
Our event fields unlock the full advantages of typical light fields—like post-capture refocusing and depth estimation—now supercharged for high-speed and high-dynamic range scenes.
% , bringing dynamic moments into sharp, vivid focus!
This novel light-sensing paradigm opens doors to new applications in photography, robotics, and AR/VR, and presents fresh challenges in rendering and machine learning.
\end{abstract}    
\vspace{-10pt}
\section{Introduction}
\label{sec:intro}

Capturing high-speed visual phenomena in real-time is essential for advancing science and engineering. 
Traditional high-speed frame cameras like Phantom, Photron, and IDT can achieve impressive frame rates (up to 100,000 fps) but are bulky, have finite capture duration, and are expensive (exceeding \$100,000), restricting accessibility. 
In contrast, event cameras offer a more affordable and efficient solution with high temporal resolution (down to $1 \mu$s) and continuous data acquisition, suitable for real-time applications. 
Unlike traditional cameras, which require substantial bandwidth (e.g., 100,000 fps, HD capture requires 100 GBps), event cameras capture only changes in the scene using asynchronous pixel readouts, reducing bandwidth demands and enabling continuous streaming. 
This approach has recently been harnessed for high-speed applications, including optical flow, object tracking, high-dynamic range imaging, structured light scanning, and photometric stereo~\cite{gallego2020event,chakravarthi2024recent}.

In this work, we extend event camera capabilities by introducing angular capture, allowing for post-capture manipulation of high-speed video---a previously untapped functionality.  Capturing angular radiance information is known as light field imaging~\cite{levoy1996light,levoy2006light,wu2017light}, enabling features such as post-capture refocusing and depth estimation. We propose ``Event Fields'' that record radiance \textit{derivatives} across angular and temporal dimensions, extending light field imaging principles to high-speed, high-dynamic-range scenarios (see \cref{fig:sec01_01_teaser}). To capture these Event Fields, we introduce two complementary hardware systems: a kaleidoscope-based design and a galvanometer-based design, each with unique advantages for this purpose.

The kaleidoscope design (see \cref{fig:sec05_02_hardware_design_kaleidoscope}), inspired by Manakov \etal\cite{manakov2013reconfigurable}, captures angle-resolved time derivatives by placing a rectangular kaleidoscope in the optical path. This approach allows for capturing angular information affordably and simply at $3 \times 3$ views.
% and high-frequency temporal events. 
The galvanometer design (see \cref{fig:sec05_01_hardware_design_galvo}) uses a galvanometer positioned in the optical path to capture angular derivatives directly. By controlling the mirrors at 250 Hz, the incoming rays are shifted along a curve, enabling the capture of both angular and temporal derivatives. As angular derivatives are sparse \cite{zhao2024survey,jin2020learning}, this design does not significantly increase bandwidth requirements. \Cref{tab:design_comparison} compares the relative advantages of these two hardware approaches.

To systematically evaluate the relative merits of the hardware designs, we build a Blender plugin to render event-fields with physics based light simulation. 
We also build hardware imaging prototypes with kaleidoscope and galvanometer, and algorithms to demonstrate new imaging capabilities in challenging environments. These include post-capture refocusing on dynamic scenes (e.g., a rotating fan at 480 RPM), focus tracking of nerf darts traveling at 20 m/s, high-speed depth reconstruction, and high-resolution, high dynamic range light fields.
% We also build two hardware prototypes using kaleidoscope design and Galvometer design. 
% We design algorithms for new imaging capabilities and with the hardware prototypes show that we can post-capture refocus on dynamic scenes (rotating fan at 480 RPM), focus track nerf bullets traveling at around 20~m/s, high-speed depth reconstruction, and high-resolution, high dynamic range light fields. 

\noindent \textbf{Contributions.} To state explicitly, the manuscript makes the following contributions: 
\begin{itemize}
\item Introduces \emph{event fields}; spatial and temporal multiplexing techniques to capture them.
\item Builds a physics-based simulator and two hardware prototypes to systematically evaluate the merits and limitations of the multiplexing techniques. 
% \item Introduces ``event fields'' and two hardware prototypes, systematically evaluating the merits and limitations of each design
% % \item Introduces ``event fields" and two hardware designs to capture them.
% \item Builds a physics-based simulator to evaluate the performance of the designs. 
% % \item A hardware prototype for each of the designs to systematically evaluate the merits and limitations. 
\item New applications and related algorithms, including slow motion refocus (SloMoRF), true-depth estimation, and high dynamic range light field imaging. 
\item Open source simulator, data, and processing codes.\footnote{We shared them as supplementary due to double-blind restrictions and will open-source them in the final version.}
\end{itemize}

% %%%%%%%%%%%%%%%%%%%%%%%%

% Define custom colors
\definecolor{headerblue}{RGB}{0, 51, 102} % Dark blue for header
\definecolor{lightgray}{RGB}{245, 245, 245} % Light gray for alternate rows
\definecolor{lightblue}{RGB}{234, 242, 251} % Very light blue for alternate rows
\definecolor{good}{RGB}{234, 251, 242} % Very light green
% \definecolor{bad}{RGB}{251, 234, 242} % Very light green
\definecolor{bad}{RGB}{252, 210, 210} % Very light green

\begin{table}[t]
    \centering
    \setlength{\arrayrulewidth}{1pt} % Thicker line for table borders
    % \renewcommand{\arraystretch}{1.5} % Increase row height for readability
    
    % \begin{tabular}{
    % |>{\columncolor{headerblue}\color{black}}c|
    % >{\columncolor{headerblue}\color{black}}c|
    % >{\columncolor{headerblue}\color{black}}c|}
    %     \hline
    %     & \color{white}\textbf{K-lens} & \color{white}\textbf{Galvo} \\ \hline
    %     \rowcolor{lightgray}
    %     Derivatives & Temp. & Ang. + Temp. \\ \hline
    %     \rowcolor{lightblue}
    %     Recon. Perf. & Scene dep. & \color{green}Scene indep. \\ \hline
    %     \rowcolor{lightgray}
    %     Color & Yes & No \\ \hline
    %     \rowcolor{lightblue}        
    %     Static LF capture& No & Yes \\ \hline
    %     \rowcolor{lightgray}
    %     Dynamic LF capture& Yes & Yes \\ \hline
    %     \rowcolor{lightblue}
    %     Refocus Fidelity & Low & High \\ \hline
    %     \rowcolor{lightgray}
    %     Bandwidth Requirements & Low & High \\ \hline
    % \end{tabular}
    \begin{tabular}{l|cc}
    \toprule
         & Kaleidoscope & Galvanometer  \\
    \midrule
        Derivatives        &  Temp. & Ang. + Temp. \\ 
        Recon. Performance       & \cellcolor{bad}Scene dep. & \cellcolor{good}Scene indep. \\
        Color              & \cellcolor{good}Yes & \cellcolor{bad}No \\ 
        Static LF Capture  & \cellcolor{bad}No & \cellcolor{good}Yes \\ 
        Dynamic LF capture & \cellcolor{good}Yes & \cellcolor{good}Yes \\ 
        Refocus Fidelity   & \cellcolor{bad}Choppy & \cellcolor{good}Smooth \\ 
        Bandwidth Required & \cellcolor{good}Low & \cellcolor{bad}High \\ 
    \bottomrule
    \end{tabular}
    \caption{Relative advantages of kaleidoscope vs. galvanometer design 
    for capturing event fields.}
    % FOV is field-of-view and LF is light field.
    %Adithya: May be we should revisit the entries of the table. We can add angular resolution, temporal resolution, trades off: spatial res and temporal res for other
    \label{tab:design_comparison}
\end{table}
\section{Related Work}
\label{sec:related_work}

\noindent {\bf Light field sensing:}
One of the first approaches to light field capture~\cite{yang2002real} involves using multiple sensors to gather angular information from various perspectives. However, this setup is bulky, costly, and requires high bandwidth communication. 
An alternative is time-sequential capture, which mounts a sensor on a mechanical gantry to capture the light field of a static scene \cite{levoy1996light, 10.1145/237170.237200}. Rather than moving the sensor itself, a mirror can be used to redirect the view. Ihrke \etal\cite{ihrke2008fast} proposed using a movable planar mirror to shift the view temporally. However, this method requires individual calibration for each image, adding complexity. Inspired by their approach, we developed a galvanometer-based design, precisely controlling the mirror's movement with a signal generator. 

In addition to time-sequential capture, angular information can also be measured with spatial multiplexing. One way to achieve this is through a microlens array~\cite{ng2005light}, which has been widely adopted in commercial light field cameras like Raytrix \cite{raytrix20173d} and Lytro \cite{lytro2017}. 
% Fusion with a high-resolution RGB camera~\cite{boominathan2014improving} was used to compensate for the loss in spatial resolution and enhance refocusing capability. 
While effective, microlens arrays are challenging to manufacture and install. Taguchi \etal\cite{taguchi2010axial} suggested using a mirror ball to expand the camera’s field of view, though this setup results in the camera being visible in the mirror ball's reflection. Alternatively, Manakov \etal\cite{manakov2013reconfigurable} introduced a kaleidoscope-like mirror array that can be attached to a standard camera without modification. 
Inspired by this design, we adapted it to convert a conventional event camera into a device capable of capturing the event field.

\noindent {\bf Event camera designs:}
From the first neuromorphic sensor by Mahowald \etal~\cite{mahowald1994silicon} to the recent Sony IMX636, event sensing saw exponential growth.
We briefly review designs most relevant to our manuscript and refer the readers to recent surveys~\cite{gallego2020event,chakravarthi2024recent}.
% For high-speed 3D tracking on sparse scenes, Shah et al. \cite{shah2024codedevents} placed phase masks in the aperture plane of the event cameras with a depth-varying point spread function (PSF). 
Event cameras are regularly used in robotics and autonomous navigation for optical flow computation~\cite{benosman2013event,gehrig2021raft, wan2022learning}, feature tracking~\cite{tedaldi2016feature}, and motion segmentation~\cite{mitrokhin2020learning, zhou2021event}, but event cameras cannot detect edges in the direction of the camera motion. 
Inspired by microsaccades in human eye movement, He \etal\cite{he2024microsaccade} placed a rotating wedge mask before the event camera aperture to capture edges in all directions. 
Similar to them, our galvanometer design detects edges oriented in all directions, independent of camera and object motion, as shown in \cref{fig:sec06_09_horizontal_vertical}. 
Our galvanometer design also finds inspiration from the designs of Mugalikar \etal\cite{muglikar2023event}, who used a rotating polarizer in the aperture plane to capture high-fidelity polarization information for shape-from-polarization and Yu \etal\cite{yu2024eventps}, who used a rotating light source and event camera for fast photometric stereo.

Habuchi \etal \cite{habuchi2024time} captured light fields for static scenes by placing an LCOS device in the aperture plane of event cameras and using CNN to recover the light field. While the technique is quite novel, it is limited to static scenes due to the LCOS response time (\SI{22}{\milli\second}). Guo \etal \cite{guo2024eventlfm} placed a microlens array (MLA) on the event sensor to capture the light field for dynamic scenes and showed microscopic imaging applications with their prototype. We built our prototypes for macroscopic applications, and our kaleidoscope design is functionally same as MLA design. However, the galvanometer design can capture angular information even for static scenes without trading off the spatial resolution like Guo \etal~\cite{guo2024eventlfm}. 

\noindent {\bf Event processing algorithms:}
% --------------------------- Adi's suggestion ---------------------------
%HDR
%Highspeed
%Interpolation (TimeLens++)
% --------------------------- HDR, Highspeed ---------------------------
% Event cameras record pixel polarity only when brightness changes exceed a threshold, allowing for 
%
Event cameras capture scene information at higher temporal resolution (in $\mu$s) and dynamic range (140~dB) compared to standard cameras (60~dB) \cite{gallego2020event}. These features enable solutions to challenges where standard frame-based sensors struggle, such as high-speed motion estimation \cite{gallego2017event}, high frame rate reconstruction \cite{rebecq2019high}, and high dynamic range imaging \cite{rebecq2019high}.
%
% --------------------------- Interpolation (Sensor fusion with RGB) ---------------------------
Besides, event cameras and standard frame-based cameras are complementary; event cameras provide high-speed, sparse data, while standard cameras capture color and fine details. By fusing these two types of sensors, algorithms like Timelens \cite{tulyakov2021time} and Timelens++ \cite{tulyakov2022time} effectively interpolate RGB frames using event information. 
We adapt these event processing algorithms for processing event field data.

\noindent {\bf Other event-based fusions:} 
% Event cameras have distinct advantages over other computational imaging and computational photography sensors. Hence, fusing event sensors with other modalities can offer the best of both imaging systems, similar to how the proposed event fields enable new imaging and vision capabilities. Below, we briefly review these techniques. 
%Sensor fusion
Apart from RGB cameras, other computational imaging systems can also be made fast by fusing with event sensors. Event-based structured light sensing techniques~\cite{muglikar2021esl, huang2021high, matsuda2015mc3d, leroux2018event,mangalore2020neuromorphic} reconstruct 3D scenes at high speed by replacing standard intensity sensors with event cameras. Event cameras are also used for foveated depth sensing~\cite{muglikar2021event} to scan moving objects at a higher spatial resolution. 
Muglikar \etal~\cite{muglikar2024event} combined single-photon avalanche diodes (SPADs) with event cameras to achieve high-speed, low-light, low-bandwidth imaging capabilities.

\section{Technical Background}
\label{sec:technical_background}

We will briefly review the image formation models for light field and event cameras that are useful in developing the mathematical models for event fields. 

%================================================
\subsection{Light Field}
\label{ssec:lf_background}
The plenoptic model of light $L(\xvec, \omegavec, \lambda, t)$ is a comprehensive representation that captures all the dimensions of light emitted by a scene~\cite{adelson1991plenoptic,mcmillan2023plenoptic}. Here, $\xvec = (x, y, z)$ represents the spatial coordinates of the observer through which the light passes; $\omegavec = (\omega_x, \omega_y)$ are the angular coordinates; $\lambda$ is the wavelength, and $t$ is the time. At a given time, the measurement of a monochromatic light is 5D function. 
Levoy and Hanrahan \cite{levoy1996light} and Gortler \etal \cite{10.1145/237170.237200} showed that the 5D representation has redundancies and have reduced it to 4D by assuming that the light field is measured in free space. Under this assumption, as the radiance of a ray remains a constant in a straight line, the static monochromatic light field is a 4D function $L(\uvec, \svec)$, where $\uvec = (u, v)$ is a point on the virtual sensor and $\svec=(s, t)$ is a point on the virtual aperture. 

%Refocusing part ?
To synthesize an image from a given light field, we integrate the light field over the aperture as:
\begin{align}
I(x, y) = \int_{\svec}L(\uvec, \svec) \mathrm{d}\svec. 
\label{eq:light_field_to_image}
\end{align}
Light field allows us to change the focus plane after imaging by computationally changing the distance between the sensor and aperture planes~\cite{levoy1996light,bishop2011light}. 
If $d_0$ is the original distance between the planes and $d$ is the distance required to focus at a given depth, the light field that focuses at $d$ will be $L_d(\uvec, \svec) = L(\uvec + (d_0/d) \svec, \svec)$. The refocused image can be synthesized by integrating angles, as shown in \cref{eq:light_field_to_image}. 

%====================================
\subsection{Event Camera}
\label{sec:tb_event_camera}
An event camera asynchronously records the brightness ($B \doteq \log(\int_{\omegavec} L d\omegavec)$) change event for each pixel independently~\cite{
mahowald1994silicon, lichtsteiner2008128,brandli2014240, posch2010qvga}. Specifically, in an ideal and noise-free scenario, a pixel $\mathbf{x} = (x_k, y_k)^T$ will be triggered at time $t_k$ as soon as the brightness change reaches a predefined threshold $C$ since the last triggered event. The polarity $p_k \in \{ +1, -1 \}$, which is the sign of the brightness change, will be recorded for each trigger. Mathematically,
\begin{align}
\Delta B(\mathbf{x}_k, t_k) &\doteq B(\mathbf{x}_k, t_k) - B(\mathbf{x}_k, t_k - \Delta t_k) = p_k C  \nonumber \\
\frac{\partial B(\mathbf{x}_k, t_k)}{\partial t} &= \frac{p_k C}{\Delta t_k} 
\Rightarrow \frac{\partial L(\mathbf{x}_k, t_k)}{\partial t} = L \frac{p_k C}{\Delta t_k} . 
\label{eq:event_camera_model}
\end{align}
Therefore, event cameras capture a high-speed temporal derivative slice ($\mathbf{x}, t_k$) of the plenoptic function. 

In the related work, we referred to several techniques that excel at extracting high-speed spatial information from sparse event streams. 
These algorithms can do so as the scene information typically changes slowly~\cite{gallego2020event}, making the temporal gradient $\frac{\partial L(\mathbf{x}_k, t_k)}{\partial t}$ inherently sparse. This redundancy also applies to angular dimension as light field is mostly smooth~\cite{zhao2024survey,jin2020learning} and hence, $\frac{\partial L(\xvec, \omegavec, \svec, \lambda, t)}{\partial \omegavec}$ and $\frac{\partial L(\xvec, \omegavec, \svec, \lambda, t)}{\partial t}$ are sparse. We exploit this sparsity to capture event fields with a single event camera in the next section.

\section{Event Field}
\label{sec:event_field}

%Adithya's text
As mentioned briefly in \cref{sec:tb_event_camera}, an event camera's photodetector measures brightness by taking the logarithm of radiance integrated over all angles. This integration results in a loss of angular information. Our objective, through event fields is to retain this angular information $\omegavec$. To achieve this goal, we need to multiplex angular information into either the spatial or temporal dimensions that the event camera captures. We advocate for and design both spatial and temporal multiplexing techniques, as they offer complementary advantages. 

\subsection{Spatial Multiplexing}
\label{sec:derivatives_time}
We encode discrete angular information ($\omega_i$) into the spatial dimension using a mapping function defined as:
\begin{equation}
    (\omega_i, \mathbf{x_s}(\omega_i)) \mapsto \mathbf{x},
\end{equation}
where $\mathbf{x_s}(\omega_i)$ represents the spatial coordinate for each measured view $(\omega_i)$. 
Let the number of discrete angular views be $\mathbf{n}$, the spatial resolution of the sensor be $\mathbf{r}$, and $\mathbf{r}$ is an integral multiple of $\mathbf{n}$. 
The specific mapping function $\mathbf{x_s}$ varies based on the hardware design. \
The two commonly used mapping functions are:
\begin{itemize}
    \item \textbf{Multi-sensor array~\cite{yang2002real} or kaleidoscope~\cite{manakov2013reconfigurable}}: In this design, the mapping function is $\mathbf{x_s} = \mathbf{x} \bmod (\mathbf{r/n})$.
    \item \textbf{Microlens array~\cite{ng2005light}}: In this design, the mapping function is $\mathbf{x_s} = \mathbf{x} // \mathbf{n}$.
\end{itemize}

By applying this mapping function to \Cref{eq:event_camera_model}, the measurements are:
\begin{equation}
    \frac{\partial B}{\partial t} (\underbrace{\mathbf{x_s}, \omega}_{\mathbf{x}_k}, t_k) \approx \frac{p_k C}{\Delta t_k}, 
\end{equation}
and the event camera will now measure the temporal brightness changes across both spatial and angular dimensions. 
Note that the brightness derivatives are still related to time, and we lose spatial resolution (by a factor of $n$) to improve angular resolution. 

\subsection{Temporal Multiplexing}
\label{sec:derivatives_angle}

To encode angular information in the temporal dimension, we steer the incoming rays along a periodic curve $\mathcal{C}(t)$ with period $T$. This steering could be accomplished by either moving the camera or placing steerable galvanometer in the optical path of the camera. The galvanometer can be steered fast along Lissajous curves~\cite{polavcek2019laser}; hence, we use it in all our experiments.

The temporal multiplexing does not sacrifice spatial resolution but sacrifices temporal resolution and always detects more events than spatial multiplexing. At any given time instant, the angular view will be at $\mathcal{C}(t \bmod T)$ and the temporally varying light field is $L(\mathbf{x}, \mathcal{C}(t \bmod T), t)$. For static scenes, the event camera measurement is
\begin{align}
\frac{\partial B(\mathbf{x}, \mathcal{C}(t))}{\partial t} = \frac{\partial B(\mathbf{x}, \omega)}{\partial \omega} \left. \frac{\partial \mathcal{C}(t)}{\partial t} \right|_{t=t_k} \approx \frac{p_k C}{\Delta t_k},
\end{align}
Therefore, temporal multiplexing technique measures the angular derivatives of the light field $\frac{\partial L(\mathbf{x}, \omega)}{\partial \omega}$ along the scanning curve. 

While we restricted the scanning curves to be Lissajous curves, specifically circles in most of the results, in principle, the scanning curve can be any periodic curve, including space-filling curves~\cite{sagan2012space,gupta2018optimal}. 
Space-filling curves result in dense angular light-field capture at the cost of loosing temporal information. However, typically light field is highly redundant and smooth and a circular scan captures most of the light-field diversity as we show in the results.

\section{Simulator and Hardware Design}

To validate our approach, we built a physics-based simulator and two hardware prototypes, kaleidoscope for spatial multiplexing and galvanometer for temporal multiplexing. 
The simulator allows us to precisely control the scene and design parameters, enabling an accurate and fair comparison between the two multiplexing techniques.
Hardware prototypes show the real-world feasibility of our designs as well as new capabilities. 

% In our hardware design, we realized spatial multiplexing with kaleidoscope design and 

% . The first, a kaleidoscope design, captures multi-view derivatives with respect to time, while the second, a galvanometer-based design, captures derivatives with respect to angle along predefined curves. We discuss the optical configurations and ray paths of each design, explaining how these setups facilitate the capture of different derivatives.

%-------------------------------------------------------------------------
% Brief deisgn of a simulator (mostly from the grant)
%-------------------------------------------------------------------------
\begin{figure}[t]
    \centering
    \includegraphics[width=\linewidth]{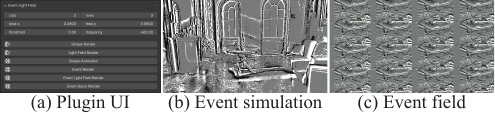}
    \caption{
    \textbf{Blender plugin} for physics-based event field rendering. 
    % \textbf{Blender simulator.}  
    % We developed a Blender plugin for physics-based rendering of spatial and temporal multiplexing designs. We compare the advantages and limitations of the two designs fairly using this simulator. 
    % This simple simulator is calculated completely according to the definition of the event. It is unbiased.
    }
    \label{fig:sec_05_03_blender_simulator}
\end{figure}

\subsection{Simulator}
We have used a straightforward approach to simulating events, which involves rendering the full plenoptic function, calculating brightness by propagating the plenoptic data through optics, and tracking the brightness difference from the previous event. A new event is triggered for each pixel when this difference exceeds a manually set threshold.

We chose Blender~\cite{blender} as our backend as it is open-source, physically based, supports GPU acceleration (Cycles), Python APIs, and brandishes a user-friendly GUI. We developed a Blender plugin to render event fields for either spatial or temporal multiplexing design, along with a simple GUI to facilitate community use. The GUI and sample simulation results are presented in \cref{fig:sec_05_03_blender_simulator}.

% Our kaleidoscope design is simulated as a $n \times n$ grid camera array, generating events for each camera. For the galvanometer design, we simulate a single camera moving in a circular curve $\mathcal{C}$ parallel to the image plane, with a frequency dependent on scene dynamics. The GUI and simulation results are presented in \cref{fig:sec_05_03_blender_simulator}.

%-------------------------------------------------------------------------
% Kaleidoscope Design Explanation (mostly from the grant)
%-------------------------------------------------------------------------
\begin{figure}[t]
    \centering
    \includegraphics[width=\linewidth]{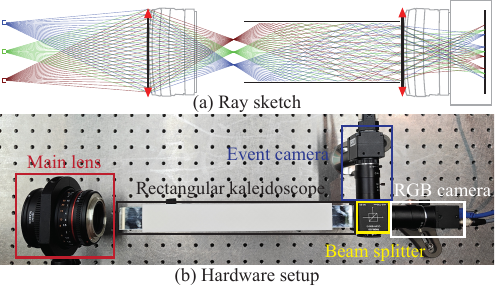}
    \caption{\textbf{Kaleidoscope design.} (a) Ray sketch illustrating the light path through the rectangular kaleidoscope system, where rays from different angles are directed towards the event camera. (b) Hardware setup showing the main lens, rectangular kaleidoscope, beam splitter, event camera, and RGB camera. The main lens focuses the scene onto the kaleidoscope, which splits the view into multiple angles. The beam splitter directs part of the light to the RGB camera for standard imaging and part to the event camera to capture high-frequency events with angular differentiation.}
    \label{fig:sec05_02_hardware_design_kaleidoscope}
\end{figure}

\subsection{Kaleidoscope for Spatial Multiplexing}
A kaleidoscope is a common optical instrument composed of multiple first-surface mirrors. Manakov et al.~\cite{manakov2013reconfigurable} introduced this optical device as an add-on for sensor fusion. As shown in \cref{fig:sec05_02_hardware_design_kaleidoscope}, this kaleidoscope is a rectangular design made of four mirrors. 
The lens at the front forms an image at the entrance of the kaleidoscope. 
After multiple reflections within the kaleidoscope, the rear camera captures the image, resulting in several flipped images, each with a different virtual camera location. Therefore, the light field information is spatially multiplexed onto the camera sensor.

% This design corresponds to \cref{sec:dervitaves_time}and enables capturing event field information with respect to time across multiple views. By using an event camera instead of a standard camera, the brightness changes across different angles are spatially multiplexed onto the event sensor, providing the event field information we require.

%-------------------------------------------------------------------------
% Galvanometer Design Explaination (mostly from the grant)
%-------------------------------------------------------------------------
\begin{figure}[t]
    \centering
    \includegraphics[width=\linewidth]{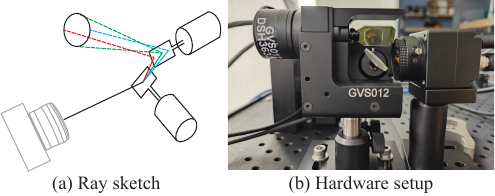}
    \caption{\textbf{Galvanometer design.} (a) Ray sketch illustrating the angular separation achieved by controlling mirror angles in the galvanometer setup. (b) Hardware setup showing the galvanometer mirror system integrated with the event camera. This design allows for capturing angular derivative events by dynamically controlling the mirror position.}
    \label{fig:sec05_01_hardware_design_galvo}
\end{figure}

\subsection{Galvanometer for Temporal Multiplexing}
While the kaleidoscope design offers a simple solution without moving parts, it has two key limitations. First, although it provides angularly separated events, the events still represent the derivative of light intensity over time. Second, the spatial resolution of current event cameras is not very high (1 megapixel), and this resolution would further decrease with spatial multiplexing. Temporal multiplexing offers an alternate approach. 

For temporal multiplexing, we place a galvanometer, as shown in \cref{fig:sec05_01_hardware_design_galvo}, in the optical path of the event camera. 
A galvanometer consists of two mirrors that can be precisely adjusted at high speed (Eg. GVS012 from Thorlabs can scan a thousand Lissajous curves per second). 
Moving mirrors allow us to rotate the camera virtually at a high speed and capture the event field.
In this design, the captured information represents the derivative of light intensity with respect to angle, rather than time, as we discussed in \cref{sec:derivatives_angle}.

\begin{figure*}[ht]
  \centering
  \includegraphics[width=\linewidth]{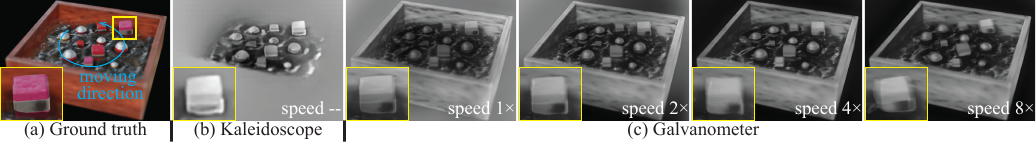}
  \caption{\textbf{(Simulation) comparison of spatial (kaleidoscope) and temporal (galvanometer) multiplexing techniques}. (a) A simulated scene with swirling water and moving objects, with a stationary box. The swirl direction is marked in blue, and one object is highlighted for detailed comparison. (b) Post-refocused image using the kaleidoscope design, showing reduced spatial resolution. However, the image quality is independent of scene motion. The static wooden box is not reconstructed. (c) Post-refocused images using the galvanometer-based system at different object movement speeds (1$\times$, 2$\times$, 4$\times$, and 8$\times$) while maintaining a constant scanning speed. 1$\times$ corresponds to about 2.5 pixels per scan for the highlighted object
  % hence, 8$\times$ is 20 pixels per scan. 
  As the scene motion increases, the highlighted object, unlike in the kaleidoscopic design, suffers from motion blur. The static wooden boxes are perfectly reconstructed, unlike the kaleidoscopic design.
  % As the scene motion increases, the sharpness in the refocused area decreases due to motion blur. The static wooden boxes are perfectly reconstructed unlike the kaleidoscopic design.
  }
  \label{fig:sec06_03_simulator_compare}
\end{figure*}
%================================================================
%================================================================
\begin{figure}[t]
    \centering
    \includegraphics[width=\linewidth]{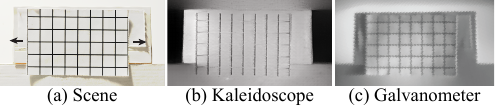}
    \caption{\textbf{(Experimental) comparison of multiplexing techniques.} (a) Reference scene with a grid pattern. (b) When the grid moves horizontally, the kaleidoscope design cannot capture horizontal lines as intensity does not change along this axis. (c) Using a galvanometer that scans in a circular motion ensures complete capture, avoiding information loss in the angular domain.} 
    % \caption{\textbf{Missing lines in horizontal movement.} (a) Scene reference of the captured grid pattern. (b) When moving the grid horizontally, the k-lens design misses horizontal lines because no events are triggered along them. (c) Using a galvanometer that scans in a circular motion ensures complete capture, avoiding information loss in the angular domain.} 
    \label{fig:sec06_09_horizontal_vertical}
\end{figure}
%================================================================
%================================================================
\begin{figure}[t]
    \centering
    \includegraphics[width=\linewidth]{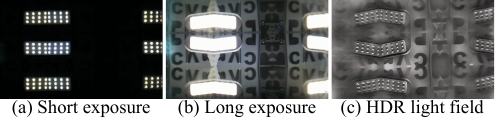}
    \caption{\textbf{High dynamic range (HDR) light field.} A high dynamic range scene is captured with a standard RGB camera using (a) short exposure and (b) long exposure, and both fail to capture the full dynamic range. In contrast, (c) the HDR light field reconstructed from the event field effectively captures the complete dynamic range of the scene. 
    % \pchakra{wouldn't it be better to do short, long and then HDR?} \apedired{I concur. Quinton, please fix it.}
    } 
    \label{fig:sec06_08_hdr}
\end{figure}
%================================================================

%============================================================================================
\begin{figure*}[t]
    \centering
    \includegraphics[width=\linewidth]{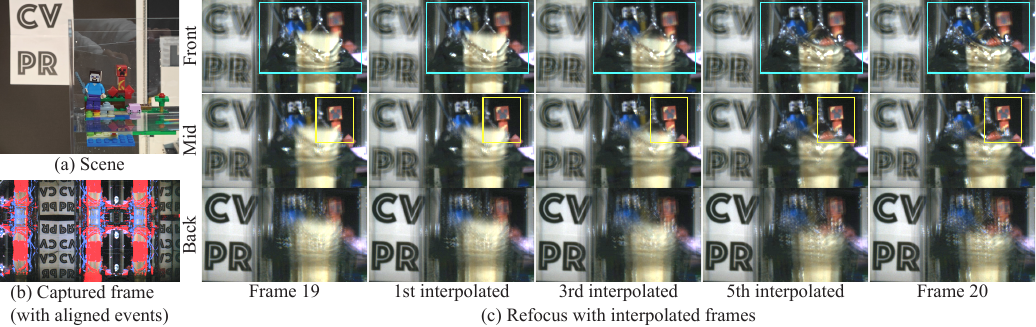}
    \caption{\textbf{SlowMoRF: Slow motion refocusing with kaleidoscope and sensor fusion.} 
    (a) Original setup with Lego pieces floating in a water tank as a wooden brick is thrown in, with CVPR in the background, (b) Captured raw frames with aligned events using kaleidoscope design, and (c) Interpolated refocused frames at three focal planes (back, mid, and front). Each column represents scene progression with interpolated frames in between. Between the frames, the fall of the wooden block and changes in water splash and caustics can be observed in the front refocus. Mid and back refocus show sharp Lego toys and CVPR. This setup achieves high-speed (720 fps) refocusing with enhanced clarity across focal depths. The supplementary contains the full video. 
    % \apedired{Right and Left are confusing and why are they in reverse order. I believe frame=10 (replace with accurate number) and frame=11 should be fine. }
    } 
    \label{fig:sec06_05_klens_refocus}
\end{figure*}
%============================================================================================

%============================================================================================
\begin{figure}[t]
    \centering
    \includegraphics[width=\linewidth]{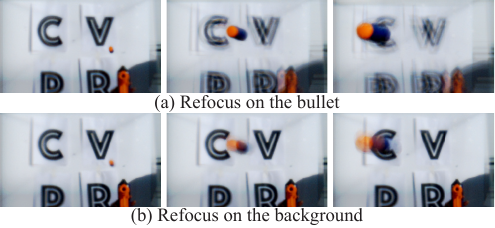}
    \caption{\textbf{High-speed refocusing on a toy dart.} The original video was captured at 120 fps, interpolating seven frames between each pair of original frames, resulting in a 960 fps high-speed light field video. The images shown here are all the middle (the fourth) interpolated frames, demonstrating refocusing on two planes: (a) the moving dart and (b) the background. This refocusing capability enables sharp detail capture of both fast-moving objects and static background elements in distinct focal planes.} 
    \label{fig:sec06_06_gun_refocus}
\end{figure}
%============================================================================================

%================================================================
\begin{figure}[t]
    \centering
    \includegraphics[width=\linewidth]{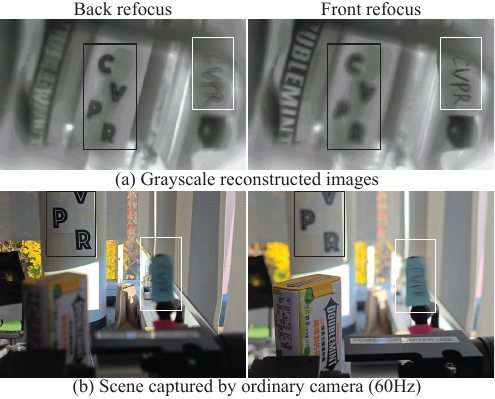}
    \caption{\textbf{Dynamic scene refocusing.} Using our galvanometer design at 250 Hz, we capture a scene with a rotating fan, marked with ``CVPR'' on the right, spinning at 300 rpm (0.2 seconds per rotation). Using E2VID, we reconstruct 40 light field frames, and (a) Using post-capture refocusing algorithm and the prior knowledge of the location of the scanned mirror, we focus on the background (left) or the rotating fan in the foreground (right). 
    % (b) Intensity values are mapped to varying colors for better visualization. 
    (b) The scene is captured by an ordinary camera at 60Hz while focusing on background (left) and foreground (right) from a slightly different viewpoint). The ordinary camera cannot provide refocus ability, and the slower shutter speed causes motion blur on the fan. 
    % \apedired{Show at least fan in real images. Can you use same colored boxes for all three rows. I understand that they might not work but find colors that would work.} 
    % \vboomi{Not a fan of (b) ``heat map''. Instead, can use space to show zoom-in of boxes in (a).}
    } 
    \label{fig:sec06_02_dynamic_scene_refocus}
\end{figure}
%================================================================

%================================================================
\begin{figure*}
    \centering
    \includegraphics[width=\linewidth]{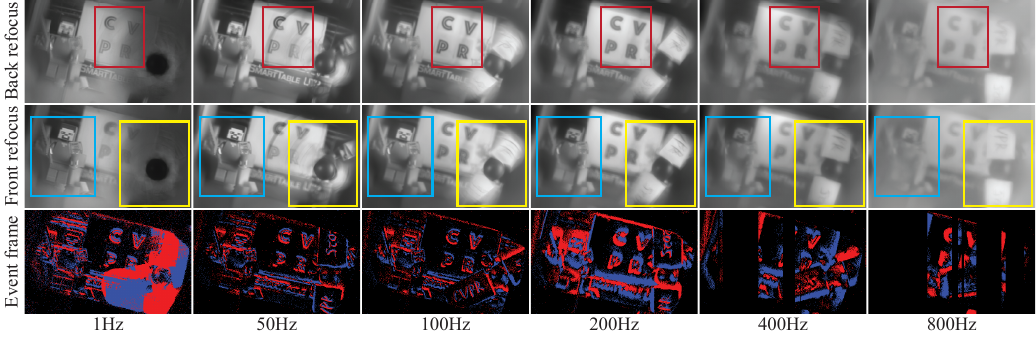}
    \caption{\textbf{Effect of galvo scan frequency for refocusing accuracy.} The scene consists of two static objects positioned in the foreground and background, along with a fan (highlighted in yellow boxes) in the foreground rotating at 480 RPM. The static objects (highlighted in red and blue boxes) remain sharp up to a scanning frequency of 200 Hz but become increasingly blurry as the scan speed increases. This counter intuitive phenomenon occurs as event camera reaches its maximum bandwidth capacity, resulting in random event loss, as seen in the \textit{Event frame} row. However, the text on the rotating fan (highlighted in yellow boxes) becomes clearer at higher scan speeds due to reduced motion blur.} 
    \label{fig:sec06_04_scan_frequency_test}
\end{figure*}
%================================================================

\section{Results}
%-------------------------------------------------------------------------
% A general introduction
%-------------------------------------------------------------------------
% Based on our two hardware designs, we present a range of results that showcase the unique strengths and limitations of each approach. First, we compare both designs through simulations, highlighting that each is essential in different contexts. The kaleidoscope design excels at capturing dynamic scenes with thousands of frames but suffers from low resolution and cannot reconstruct static scenes, as no events are generated when there is no motion. In contrast, the galvanometer design achieves full-resolution recovery of both static and dynamic scenes by capturing angular derivatives. However, unlike the kaleidoscope, the galvanometer requires time to complete a full scan, leading to potential motion blur if objects move significantly during scanning.

We evaluate the proposed techniques on various scenes. 
We first summarize the observations and then provide details. 
% on various scenes with simulations and real-world measurements. 
% We present results from our two hardware designs, highlighting their unique strengths and limitations. 

\textit{Simulations} demonstrate that each design is optimal in different contexts.
The kaleidoscope (spatial multiplexing) excels in capturing dynamic scenes with high frame rates but suffers from lower spatial resolution and is ineffective for static scenes, as no events are generated without motion.
The galvanometer (temporal multiplexing) captures both static and dynamic scenes at full spatial resolution by recording angular derivatives but introduces motion blur if the object motion is significant.

% The galvanometer (temporal multiplexing) captures both static and dynamic scenes at full spatial resolution by recording angular derivatives, though it requires a full scan cycle, introducing potential motion blur if objects move significantly during scanning.

% We then extend our designs to real-world applications, demonstrating post-refocusing capabilities. With the galvanometer setup, we perform both static and dynamic scene refocusing, {\color{warning}capturing high-resolution details.} For the kaleidoscope design, we achieve sensor fusion with an RGB camera, enabling colorful, high-speed dynamic scene refocusing. These results underscore the complementary nature of each design: the kaleidoscope’s strength in high-speed dynamic capture and the galvanometer’s high-resolution angular detail recovery.

\textit{Real-world applications} demonstrate each design's post-capture refocusing capabilities.
Specifically, with the galvanometer setup, we show high-resolution refocusing of static and dynamic parts of the scene. 
With the kaleidoscope setup, we perform sensor fusion with an RGB camera, providing vibrant, high-speed refocusing for dynamic scenes.
These results underscore the complementary nature of the two designs: the kaleidoscope for high-speed dynamic capture and the galvanometer for detailed, high spatial-angular light field recovery.

In the following sections, we compare our proposed hardware designs (see \Cref{subsec:k-g-lens}), followed by applications for each setup. Specifically, we explore HDR light fields (\Cref{subsec:hdr-lf}) and slow-motion refocusing (\Cref{sec:refocusing_klens}) using the kaleidoscope configuration, as well as high-speed post-capture refocusing (\Cref{sec:refocusing_galvo}) and instant true depth estimation (\Cref{subsec:instant-depth}) using the galvanometer-based design.

% Below, we will explain the results in more detail. 

% \subsection{One same scene with two different setups}
% \subsection{Comparison of kaleidoscopic and galvanometer designs}

\subsection{Kaleidoscope vs Galvanometer Designs}
\label{subsec:k-g-lens}

In practical applications, differences in optical properties---such as focal lengths---make it challenging to compare two cameras under identical conditions. The simulator, however, allows us to control these variables, enabling a fair comparison of both designs.
We, therefore, used Blender to create a scene containing swirling water and moving objects within a stationary box (see \cref{fig:sec06_03_simulator_compare}) to assess the strengths and limitations of each design.

The first significant difference is that the galvanometer design can fully reconstruct scene information by capturing angular derivatives, whereas the kaleidoscope design can only capture moving objects, as no events are generated in static regions. 
The stationary wooden box is clearly visible in the galvanometer design but is absent in the kaleidoscope design. 
We also observe this in a real experiment (\cref{fig:sec06_09_horizontal_vertical}), where a rectangular mesh is translated horizontally, and the kaleidoscope fails to reconstruct horizontal lines.
% Additionally, we conducted an experiment with our real-world setup, as shown in \cref{fig:sec06_09_horizontal_vertical}. The kaleidoscope design failed to reconstruct horizontal lines when the grid was moved in the horizontal direction.

A second key difference is that the galvanometer design multiplexes the event field into the time domain, which makes it susceptible to motion blur when objects move rapidly. This effect is visible in the zoomed insets in \cref{fig:sec06_03_simulator_compare}. 
The kaleidoscope design, by contrast, avoids motion blur but instead suffers from lower spatial resolution.
Both designs, however, are capable of capturing and refocusing dynamic moving objects.

%-------------------------------------------------------------------------
 % High dynamic light field (k-lens design)
%-------------------------------------------------------------------------
\subsection{HDR Light Fields}
\label{subsec:hdr-lf}
% \pchakra{because we introduced the term event fields, we should use it. Also, we use K-lens, k-lens and kaleidoscope. should be consistent}
Event fields are not only high-speed but also high-dynamic range light fields. 
We demonstrate this in \cref{fig:sec06_08_hdr}, where the scene containing bright LED light in the foreground and diffuse texture in the background is captured using both an RGB camera and the event camera through a kaleidoscope. 
The event fields have a large dynamic range and capture the spatial structure of both LEDs and the background texture, whereas the RGB camera either captures the light sources (short exposure) or the background (long exposure). 
% We compare the HDR event fields with standard light fields captured by intensity sensor
% Using our K-lens design, we compared the event sensor to a standard RGB sensor.
% Leveraging the high dynamic range capability of the event camera, our system can generate HDR light fields effectively. To demonstrate this, we positioned a bright LED light in the foreground and a dark board in the background. 
% Using our k-lens design, we conducted a fair comparison between an RGB sensor and an event sensor. 
% As shown in \cref{fig:sec06_08_hdr}, the event camera's reconstructed light field successfully captures both the bright LED beads and the dark background in a single exposure, whereas 
% the RGB camera fails to do so.
% \pchakra{is comparing simultaneous capture of events and single exposure of rgb justified?}
% \apedired{I didn't understand the question. I assume the total exposure duration is same. Does that answer?}

% highlight that the reconstructed light field captures both the bright LED beads and the dark background simultaneously, whereas the standard RGB camera struggles to capture both elements within a single exposure.

%-------------------------------------------------------------------------
 % K-lens + Timelens + pattern matching
%-------------------------------------------------------------------------
% \subsection{Refocusing with K-lens Sensor Fusion}
\subsection{SloMoRF: Slow Motion Refocusing}
\label{sec:refocusing_klens}

We augment the kaleidoscope design with a co-located RGB camera 
% (running at 120 fps) 
to capture and reconstruct slow-motion RGB frames with dynamic refocusing capability, called SloMoRF.
% To enhance the capabilities of the kaleidoscope design, which only detects moving objects, we co-locate it with an RGB camera operating at 120 fps and do sensor fusion. 
% \pchakra{is this fps correct? wrote this based on later data}. 
% Since the kaleidoscope design only detects moving objects in the scene, we incorporate sensor fusion to extend its capabilities. Specifically, we combine data from an event camera and an ordinary RGB camera. 
The hardware setup, shown in \cref{fig:sec05_02_hardware_design_kaleidoscope}, features a beam splitter for identical kaleidoscope views for RGB and event camera without interfering with the main lens. 
% This configuration produces a $3 \times 3$ light field and event field grid, multiplexed spatially.

We utilize the TimeLens algorithm \cite{tulyakov2021time} for sensor fusion\footnote{code for the updated TimeLens++ \cite{tulyakov2022time} is not publicly available}. 
Using the captured event data, we interpolate frames to generate a 6$\times$ higher frame-rate RGB video (from 120 to 720 fps).
% Specifically, starting with an RGB video of 120 fps, we interpolate 5 frames between consecutive RGB frames, boosting the effective frame rate to 720 fps. 
The interpolated results are presented in \cref{fig:sec06_05_klens_refocus}.
% The state-of-the-art methods for sensor fusion are TimeLens \cite{tulyakov2021time} and TimeLens++ \cite{tulyakov2022time}. Since the code for TimeLens++ is currently unavailable, we use the TimeLens algorithm. 

This approach allows us to generate a high-speed color light field. We treat this light field as unstructured as we found it empirically easier than calibrating the views. 
To align the images, we estimate the necessary shift at a specified patch using a basic pattern-matching algorithm. 
Note that this shift computation is \textit{only needed once} for the entire video sequence as long as the target focal plane remains fixed. 
\Cref{fig:sec06_05_klens_refocus} shows refocusing results across three focal planes. 
Additionally, the captured event field allows for focus tracking on a high-speed object moving along the depth. 
We demonstrate this capability in \cref{fig:sec06_06_gun_refocus}, where we visually compare sequence of images that focus on a fast-moving toy dart vs focusing solely on the background.
These results demonstrate the high frame rate, color, and refocusing capabilities of our kaleidoscope design when combined with the state-of-the-art sensor fusion algorithms.

% By integrating a simple object tracking algorithm in OpenCV, which provides the search template for each frame on the object we are interested in, we can follow a high-speed object moving along the depth, for instance, a flying Nerf gun bullet. In Figure \ref{fig:sec06_06_gun_refocus}, we demonstrate the capability of consistently focusing on a fast-moving bullet. We visually compared focusing on the bullet with refocusing solely on the background.

%-------------------------------------------------------------------------
 % Galvo + E2VID + (pattern matching)
%-------------------------------------------------------------------------
\subsection{High-Speed Post-capture Refocusing}
\label{sec:refocusing_galvo}

% Event fields result in high-speed, high spatio-angular, HDR light fields. 
% Post-capture refocusing is one of the fundamental applications of the light field. 
% We can achieve that using the equations described in \cref{ssec:lf_background}. 
% Post-capture refocusing is one of the fundamental applications of light field. 
% We can achieve that using the equations described in \cref{ssec:lf_background}.
% To validate its effectiveness, we conduct tests on two static scenes, each with a foreground object and background elements.
% We test the refocusing ability on a dynamic scene containing a fan rotating at 300 RPM (\cref{fig:sec06_02_dynamic_scene_refocus}) using the galvanometer setup.

In this section, we test the post-capture refocusing ability of the galvanometer setup on dynamic scenes. 
Compared to unstructured event field captured by kaleidoscope design, 
our galvanometer design produces a fully structured event field after calibration, as we precisely control the galvanometer with a signal generator to determine the angle relative to time. 
The supplementary materials provide calibration details. 
% By leveraging E2VID \cite{rebecq2019high}, we convert the event field into a grayscale light field.

Using E2VID, we generate a high-speed video at 10,000 fps (0.1 ms per frame) from the captured event field data. 
The galvanometer is operating at 250 Hz. 
Therefore, we reconstruct the light field at 250 fps and 40 views, the fastest recorded so far at a megapixel resolution. 

Using this light field and equations in \cref{ssec:lf_background}, we can now do post-capture refocusing. 
Note that even though E2VID generates noisy frames at high speed (due to low light), the noise gets averaged out when we synthesize a refocused image. 
In \cref{fig:sec06_02_dynamic_scene_refocus}, we show refocusing results on the background texture (left figure) or foreground fan (right figure) and can observe that they are noise-free. 
% As shown in the top-right corner of \cref{fig:sec06_02_dynamic_scene_refocus}, with this high-speed light-field, we can effectively focus either on the fast-moving fan or on the background with no motion blur. 
We further stress-tested our hardware setup in \cref{fig:sec06_04_scan_frequency_test} and noticed that the event camera begins to lose events due to the readout bandwidth limit when the galvanometer operates beyond 250~Hz. 

\subsection{Instant True Depth Estimation}
\label{subsec:instant-depth}
In addition to refocusing, our galvanometer design enables real-time true depth estimation. Unlike SLAM-based methods~\cite{cadena2016past}, our approach provides instant true depth estimation based on the event field.
As outlined in \cref{sec:refocusing_galvo}, our galvanometer design generates a structured light field, requiring only a single calibration step to determine the depth-to-disparity ratio.
% . Consequently, the only requirement is a single calibration step to determine the depth-to-radius ratio on the sensor.
%
We employ the ``depth from focus'' algorithm \cite{grossmann1987depth} to calculate the depth of each pixel. This involves first constructing a focal stack by refocusing the light field at various depths. For each pixel, we determine the focal stack image where the pixel appears sharpest, and the depth associated with that image is the depth of the pixel.
% , which identifies the \textit{sharpest} pixel within its neighborhood at each image in the focal stack. 
% We first construct a focal stack by focusing on each depth plane.
% By selecting the image with maximum sharpness for each pixel, we \textit{construct a focal stack} corresponding to specific depths, allowing for instant true depth estimation from the event field.

% We utilize the ``depth from focus'' algorithm\cite{grossmann1987depth}, which identifies the \textit{sharpest} pixel within its neighborhood at each image in the focal stack. 
% We first construct a focal stack by focusing on each depth plane.
% By selecting the image with maximum sharpness for each pixel, we \textit{construct a focal stack} corresponding to specific depths, allowing for instant true depth estimation from the event field.
% \todo{this sentence should be improved to make it simpler}

% We determine the focal stack of each pixel by selecting the one has maximum sharpness. Each focal stack corresponding to a specific depth enables us to obtain instantaneous true depth based on our event field.

\begin{figure}[t]
    \centering
    \includegraphics[width=\linewidth]{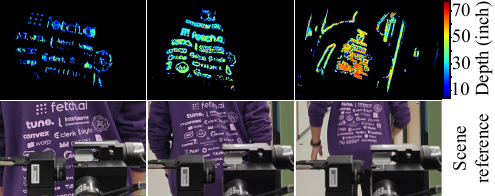}
    \caption{\textbf{Instant true depth estimation at 100 Hz.} Depth mapping of a running person, shown in three frames with color-coded depth scale (in inches), highlighting the system’s capability for rapid, accurate depth measurement in dynamic scenes.}
    % \pchakra{put an rgb reference image}} 
    \label{fig:sec06_07_depth}
\end{figure}

By adopting this approach, we obtain the instant true depth at the same frequency as the galvanometer’s scanning speed. 
As shown in \cref{fig:sec06_07_depth}, a 100 Hz galvo scan successfully captures a person running away from the camera, offering sparse yet real-time depth information of the letters on their shirt at a rate of 100 fps.

\section{Discussion and Future Work}

In this manuscript, we introduce ``event field", a new approach for capturing high-speed, high resolution, HDR light fields using two optical configurations—kaleidoscope and galvanometer—combined with an event camera. Our kaleidoscope design enables capturing temporal derivatives for multiple views, while the galvanometer design captures angular derivatives along curves. We validate both designs through simulations and real-world data, demonstrating their unique advantages.

Our experiments showcase several potential applications. Most notably, our setup enables high-speed refocusing on fast-moving objects, real-time true depth estimation, and high-dynamic-range light field capture. Supplementary PDF and video provides further visualizations of these results across additional scenes.

While our approach shows several new possibilities, it also creates new research problems in the graphics, imaging, and vision fields. 
Our current brute-force simulation of the event field could benefit from optimal importance sampling techniques that improve the rendering speed by orders of magnitude similar to previous importance techniques for computational cameras~\cite{jarabo2014framework, kim2023doppler, liu2022temporally, pediredla2020path}. 
Additionally, as shown in \cref{fig:sec06_04_scan_frequency_test}, static parts of the scene do not require high bandwidth. By implementing adaptive scanning (e.g., slower scans for static parts, faster scans for dynamic areas) with a MEMS array, we can enhance bandwidth efficiency. 
We currently use pre-trained E2VID and Timelens models, which are optimized for time-sequential event streams. Generating sufficient event field data, possibly through an optimal simulator and training models specifically for our event fields, would yield a significant performance boost.

% {\color{warning} We put all our results (and even more scenes) in supplementary as videos. Make sure to check it out to get a better visualization and understanding.}

% We do the justification for both design and shows that both of them have advantages.....

% We did the post refocusing application........

% Better sampling for the rendering algorithm to speed the simulation up. 

% Many other application can be done such as monocular depth, high speed 3D optical flow, and specific reconstruction algorithm based on event field.......

% Limitations: Galvanometer is a fixed aperture (only on curve $\mathcal{C}$)......

% if we could scan different parts of the scene adaptively at different velocities (static parts at lower velocity and faster parts at higher speed), using a MEMS array (DMD ?), we could better use device bandwidth.
% \input{sec/8_acknowledgment}
\clearpage
{
    \small
    \bibliographystyle{ieeenat_fullname}
    \bibliography{main}

\begin{thebibliography}{55}
\providecommand{\natexlab}[1]{#1}
\providecommand{\url}[1]{\texttt{#1}}
\expandafter\ifx\csname urlstyle\endcsname\relax
  \providecommand{\doi}[1]{doi: #1}\else
  \providecommand{\doi}{doi: \begingroup \urlstyle{rm}\Url}\fi

\bibitem[Adelson et~al.(1991)Adelson, Bergen, et~al.]{adelson1991plenoptic}
Edward~H Adelson, James~R Bergen, et~al.
\newblock \emph{The plenoptic function and the elements of early vision}.
\newblock Vision and Modeling Group, Media Laboratory, Massachusetts Institute of~…, 1991.

\bibitem[Anonymous(2024)]{eventfield2024}
Anonymous.
\newblock Event field.
\newblock \url{https://anonymous.4open.science/r/Event-Field-B5F3/}, 2024.

\bibitem[Benosman et~al.(2013)Benosman, Clercq, Lagorce, Ieng, and Bartolozzi]{benosman2013event}
Ryad Benosman, Charles Clercq, Xavier Lagorce, Sio-Hoi Ieng, and Chiara Bartolozzi.
\newblock Event-based visual flow.
\newblock \emph{IEEE transactions on neural networks and learning systems}, 25\penalty0 (2):\penalty0 407--417, 2013.

\bibitem[Bishop and Favaro(2011)]{bishop2011light}
Tom~E Bishop and Paolo Favaro.
\newblock The light field camera: Extended depth of field, aliasing, and superresolution.
\newblock \emph{IEEE transactions on pattern analysis and machine intelligence}, 34\penalty0 (5):\penalty0 972--986, 2011.

\bibitem[Brandli et~al.(2014)Brandli, Berner, Yang, Liu, and Delbruck]{brandli2014240}
Christian Brandli, Raphael Berner, Minhao Yang, Shih-Chii Liu, and Tobi Delbruck.
\newblock A 240$\times$ 180 130 db 3 $\mu$s latency global shutter spatiotemporal vision sensor.
\newblock \emph{IEEE Journal of Solid-State Circuits}, 49\penalty0 (10):\penalty0 2333--2341, 2014.

\bibitem[Cadena et~al.(2016)Cadena, Carlone, Carrillo, Latif, Scaramuzza, Neira, Reid, and Leonard]{cadena2016past}
Cesar Cadena, Luca Carlone, Henry Carrillo, Yasir Latif, Davide Scaramuzza, Jos{\'e} Neira, Ian Reid, and John~J Leonard.
\newblock Past, present, and future of simultaneous localization and mapping: Toward the robust-perception age.
\newblock \emph{IEEE Transactions on robotics}, 32\penalty0 (6):\penalty0 1309--1332, 2016.

\bibitem[Chakravarthi et~al.(2024)Chakravarthi, Verma, Daniilidis, Fermuller, and Yang]{chakravarthi2024recent}
Bharatesh Chakravarthi, Aayush~Atul Verma, Kostas Daniilidis, Cornelia Fermuller, and Yezhou Yang.
\newblock Recent event camera innovations: A survey.
\newblock \emph{arXiv preprint arXiv:2408.13627}, 2024.

\bibitem[Community(2024)]{blender}
Blender~Online Community.
\newblock Blender - a 3d modelling and rendering package, 2024.

\bibitem[Gallego et~al.(2017)Gallego, Lund, Mueggler, Rebecq, Delbruck, and Scaramuzza]{gallego2017event}
Guillermo Gallego, Jon~EA Lund, Elias Mueggler, Henri Rebecq, Tobi Delbruck, and Davide Scaramuzza.
\newblock Event-based, 6-dof camera tracking from photometric depth maps.
\newblock \emph{IEEE transactions on pattern analysis and machine intelligence}, 40\penalty0 (10):\penalty0 2402--2412, 2017.

\bibitem[Gallego et~al.(2020)Gallego, Delbr{\"u}ck, Orchard, Bartolozzi, Taba, Censi, Leutenegger, Davison, Conradt, Daniilidis, et~al.]{gallego2020event}
Guillermo Gallego, Tobi Delbr{\"u}ck, Garrick Orchard, Chiara Bartolozzi, Brian Taba, Andrea Censi, Stefan Leutenegger, Andrew~J Davison, J{\"o}rg Conradt, Kostas Daniilidis, et~al.
\newblock Event-based vision: A survey.
\newblock \emph{IEEE transactions on pattern analysis and machine intelligence}, 44\penalty0 (1):\penalty0 154--180, 2020.

\bibitem[Gehrig et~al.(2021)Gehrig, Millh{\"a}usler, Gehrig, and Scaramuzza]{gehrig2021raft}
Mathias Gehrig, Mario Millh{\"a}usler, Daniel Gehrig, and Davide Scaramuzza.
\newblock E-raft: Dense optical flow from event cameras.
\newblock In \emph{2021 International Conference on 3D Vision (3DV)}, pages 197--206. IEEE, 2021.

\bibitem[Gortler et~al.(1996)Gortler, Grzeszczuk, Szeliski, and Cohen]{10.1145/237170.237200}
Steven~J. Gortler, Radek Grzeszczuk, Richard Szeliski, and Michael~F. Cohen.
\newblock The lumigraph.
\newblock In \emph{Proceedings of the 23rd Annual Conference on Computer Graphics and Interactive Techniques}, page 43–54, New York, NY, USA, 1996. Association for Computing Machinery.

\bibitem[Grossmann(1987)]{grossmann1987depth}
Paul Grossmann.
\newblock Depth from focus.
\newblock \emph{Pattern recognition letters}, 5\penalty0 (1):\penalty0 63--69, 1987.

\bibitem[Guo et~al.(2024)Guo, Yang, Chang, Hu, Greene, Gabel, You, and Tian]{guo2024eventlfm}
Ruipeng Guo, Qianwan Yang, Andrew~S Chang, Guorong Hu, Joseph Greene, Christopher~V Gabel, Sixian You, and Lei Tian.
\newblock Eventlfm: Event camera integrated fourier light field microscopy for ultrafast 3d imaging.
\newblock \emph{Light: Science \& Applications}, 13\penalty0 (1):\penalty0 144, 2024.

\bibitem[Gupta et~al.(2018)Gupta, Velten, Nayar, and Breitbach]{gupta2018optimal}
Mohit Gupta, Andreas Velten, Shree~K Nayar, and Eric Breitbach.
\newblock What are optimal coding functions for time-of-flight imaging?
\newblock \emph{ACM Transactions on Graphics (TOG)}, 37\penalty0 (2):\penalty0 1--18, 2018.

\bibitem[Habuchi et~al.(2024)Habuchi, Takahashi, Tsutake, Fujii, and Nagahara]{habuchi2024time}
Shuji Habuchi, Keita Takahashi, Chihiro Tsutake, Toshiaki Fujii, and Hajime Nagahara.
\newblock Time-efficient light-field acquisition using coded aperture and events.
\newblock In \emph{Proceedings of the IEEE/CVF Conference on Computer Vision and Pattern Recognition}, pages 24923--24933, 2024.

\bibitem[He et~al.(2024)He, Wang, Zhou, Chen, Singh, Li, Gao, Shen, Wang, Cao, et~al.]{he2024microsaccade}
Botao He, Ze Wang, Yuan Zhou, Jingxi Chen, Chahat~Deep Singh, Haojia Li, Yuman Gao, Shaojie Shen, Kaiwei Wang, Yanjun Cao, et~al.
\newblock Microsaccade-inspired event camera for robotics.
\newblock \emph{Science Robotics}, 9\penalty0 (90):\penalty0 eadj8124, 2024.

\bibitem[Huang et~al.(2021)Huang, Zhang, and Xiong]{huang2021high}
Xueyan Huang, Yueyi Zhang, and Zhiwei Xiong.
\newblock High-speed structured light based 3d scanning using an event camera.
\newblock \emph{Optics Express}, 29\penalty0 (22):\penalty0 35864--35876, 2021.

\bibitem[Ihrke et~al.(2008)Ihrke, Stich, Gottschlich, Magnor, and Seidel]{ihrke2008fast}
Ivo Ihrke, Timo Stich, Heiko Gottschlich, Marcus Magnor, and Hans-Peter Seidel.
\newblock Fast incident light field acquisition and rendering.
\newblock \emph{Journal of WSCG}, 2008.

\bibitem[Jarabo et~al.(2014)Jarabo, Marco, Munoz, Buisan, Jarosz, and Gutierrez]{jarabo2014framework}
Adrian Jarabo, Julio Marco, Adolfo Munoz, Raul Buisan, Wojciech Jarosz, and Diego Gutierrez.
\newblock A framework for transient rendering.
\newblock \emph{ACM Transactions on Graphics (ToG)}, 33\penalty0 (6):\penalty0 1--10, 2014.

\bibitem[Jin et~al.(2020)Jin, Hou, Yuan, and Kwong]{jin2020learning}
Jing Jin, Junhui Hou, Hui Yuan, and Sam Kwong.
\newblock Learning light field angular super-resolution via a geometry-aware network.
\newblock In \emph{Proceedings of the AAAI conference on artificial intelligence}, pages 11141--11148, 2020.

\bibitem[Kim et~al.(2023)Kim, Jarosz, Gkioulekas, and Pediredla]{kim2023doppler}
Juhyeon Kim, Wojciech Jarosz, Ioannis Gkioulekas, and Adithya Pediredla.
\newblock Doppler time-of-flight rendering.
\newblock \emph{ACM Trans. Graph.}, 42\penalty0 (6), 2023.

\bibitem[Leroux et~al.(2018)Leroux, Ieng, and Benosman]{leroux2018event}
T Leroux, S-H Ieng, and Ryad Benosman.
\newblock Event-based structured light for depth reconstruction using frequency tagged light patterns.
\newblock \emph{arXiv preprint arXiv:1811.10771}, 2018.

\bibitem[Levoy(2006)]{levoy2006light}
M. Levoy.
\newblock Light fields and computational imaging.
\newblock \emph{Computer}, 39\penalty0 (8):\penalty0 46--55, 2006.

\bibitem[Levoy and Hanrahan(1996)]{levoy1996light}
Marc Levoy and Pat Hanrahan.
\newblock Light field rendering.
\newblock In \emph{Proceedings of the 23rd Annual Conference on Computer Graphics and Interactive Techniques}, page 31–42, New York, NY, USA, 1996. Association for Computing Machinery.

\bibitem[Lichtsteiner et~al.(2008)Lichtsteiner, Posch, and Delbruck]{lichtsteiner2008128}
Patrick Lichtsteiner, Christoph Posch, and Tobi Delbruck.
\newblock A 128$\times$128 120 db 15$\mu$s latency asynchronous temporal contrast vision sensor.
\newblock \emph{IEEE journal of solid-state circuits}, 43\penalty0 (2):\penalty0 566--576, 2008.

\bibitem[Liu et~al.(2022)Liu, Jiao, and Jarosz]{liu2022temporally}
Yang Liu, Shaojie Jiao, and Wojciech Jarosz.
\newblock Temporally sliced photon primitives for time-of-flight rendering.
\newblock In \emph{Computer Graphics Forum}, pages 29--40. Wiley Online Library, 2022.

\bibitem[{Lytro}(2017)]{lytro2017}
{Lytro}.
\newblock Lytro.
\newblock \url{https://www.lytro.com/}, 2017.
\newblock [Online; accessed 6-Oct-2024].

\bibitem[Mahowald and Mahowald(1994)]{mahowald1994silicon}
Misha Mahowald and Misha Mahowald.
\newblock The silicon retina.
\newblock \emph{An Analog VLSI System for Stereoscopic Vision}, pages 4--65, 1994.

\bibitem[Manakov et~al.(2013)Manakov, Restrepo, Klehm, Hegedus, Eisemann, Seidel, and Ihrke]{manakov2013reconfigurable}
Alkhazur Manakov, John Restrepo, Oliver Klehm, Ramon Hegedus, Elmar Eisemann, Hans-Peter Seidel, and Ivo Ihrke.
\newblock A reconfigurable camera add-on for high dynamic range, multispectral, polarization, and light-field imaging.
\newblock \emph{ACM Transactions on Graphics}, 32\penalty0 (4):\penalty0 47--1, 2013.

\bibitem[Mangalore et~al.(2020)Mangalore, Seelamantula, and Thakur]{mangalore2020neuromorphic}
Ashish~Rao Mangalore, Chandra~Sekhar Seelamantula, and Chetan~Singh Thakur.
\newblock Neuromorphic fringe projection profilometry.
\newblock \emph{IEEE Signal Processing Letters}, 27:\penalty0 1510--1514, 2020.

\bibitem[Matsuda et~al.(2015)Matsuda, Cossairt, and Gupta]{matsuda2015mc3d}
Nathan Matsuda, Oliver Cossairt, and Mohit Gupta.
\newblock {Mc3d: Motion contrast 3d scanning}.
\newblock In \emph{2015 IEEE International Conference on Computational Photography (ICCP)}, pages 1--10. IEEE, 2015.

\bibitem[McMillan and Bishop(2023)]{mcmillan2023plenoptic}
Leonard McMillan and Gary Bishop.
\newblock Plenoptic modeling: An image-based rendering system.
\newblock In \emph{Seminal Graphics Papers: Pushing the Boundaries, Volume 2}, pages 433--440. Association for Computing Machinery, 2023.

\bibitem[Mitrokhin et~al.(2020)Mitrokhin, Hua, Fermuller, and Aloimonos]{mitrokhin2020learning}
Anton Mitrokhin, Zhiyuan Hua, Cornelia Fermuller, and Yiannis Aloimonos.
\newblock Learning visual motion segmentation using event surfaces.
\newblock In \emph{Proceedings of the IEEE/CVF Conference on Computer Vision and Pattern Recognition}, pages 14414--14423, 2020.

\bibitem[Muglikar et~al.(2021{\natexlab{a}})Muglikar, Gallego, and Scaramuzza]{muglikar2021esl}
Manasi Muglikar, Guillermo Gallego, and Davide Scaramuzza.
\newblock Esl: Event-based structured light.
\newblock In \emph{2021 International Conference on 3D Vision (3DV)}, pages 1165--1174. IEEE, 2021{\natexlab{a}}.

\bibitem[Muglikar et~al.(2021{\natexlab{b}})Muglikar, Moeys, and Scaramuzza]{muglikar2021event}
Manasi Muglikar, Diederik~Paul Moeys, and Davide Scaramuzza.
\newblock Event guided depth sensing.
\newblock In \emph{2021 International Conference on 3D Vision (3DV)}, pages 385--393. IEEE, 2021{\natexlab{b}}.

\bibitem[Muglikar et~al.(2023)Muglikar, Bauersfeld, Moeys, and Scaramuzza]{muglikar2023event}
Manasi Muglikar, Leonard Bauersfeld, Diederik~Paul Moeys, and Davide Scaramuzza.
\newblock Event-based shape from polarization.
\newblock In \emph{Proceedings of the IEEE/CVF Conference on Computer Vision and Pattern Recognition}, pages 1547--1556, 2023.

\bibitem[Muglikar et~al.(2024)Muglikar, Somasundaram, Dave, Charbon, Raskar, and Scaramuzza]{muglikar2024event}
Manasi Muglikar, Siddharth Somasundaram, Akshat Dave, Edoardo Charbon, Ramesh Raskar, and Davide Scaramuzza.
\newblock Event cameras meet spads for high-speed, low-bandwidth imaging.
\newblock \emph{arXiv preprint arXiv:2404.11511}, 2024.

\bibitem[Ng et~al.(2005)Ng, Levoy, Br{\'e}dif, Duval, Horowitz, and Hanrahan]{ng2005light}
Ren Ng, Marc Levoy, Mathieu Br{\'e}dif, Gene Duval, Mark Horowitz, and Pat Hanrahan.
\newblock \emph{Light field photography with a hand-held plenoptic camera}.
\newblock PhD thesis, Stanford university, 2005.

\bibitem[Pediredla et~al.(2020)Pediredla, Chalmiani, Scopelliti, Chamanzar, Narasimhan, and Gkioulekas]{pediredla2020path}
Adithya Pediredla, Yasin~Karimi Chalmiani, Matteo~Giuseppe Scopelliti, Maysamreza Chamanzar, Srinivasa Narasimhan, and Ioannis Gkioulekas.
\newblock Path tracing estimators for refractive radiative transfer.
\newblock \emph{ACM Transactions on Graphics (TOG)}, 39\penalty0 (6):\penalty0 1--15, 2020.

\bibitem[Pol{\'a}{\v{c}}ek et~al.(2019)Pol{\'a}{\v{c}}ek, Jurmanov{\'a}, and Navr{\'a}til]{polavcek2019laser}
Lubo{\v{s}} Pol{\'a}{\v{c}}ek, Jana Jurmanov{\'a}, and Zden{\v{e}}k Navr{\'a}til.
\newblock Laser galvo mirrors: perfect instrument for the demonstration of lissajous figures.
\newblock \emph{Physics Education}, 54\penalty0 (5):\penalty0 055002, 2019.

\bibitem[Posch et~al.(2010)Posch, Matolin, and Wohlgenannt]{posch2010qvga}
Christoph Posch, Daniel Matolin, and Rainer Wohlgenannt.
\newblock A qvga 143 db dynamic range frame-free pwm image sensor with lossless pixel-level video compression and time-domain cds.
\newblock \emph{IEEE Journal of Solid-State Circuits}, 46\penalty0 (1):\penalty0 259--275, 2010.

\bibitem[Raytrix(2017)]{raytrix20173d}
A Raytrix.
\newblock 3d light field camera technology, 2017.

\bibitem[Rebecq et~al.(2019)Rebecq, Ranftl, Koltun, and Scaramuzza]{rebecq2019high}
Henri Rebecq, Ren{\'e} Ranftl, Vladlen Koltun, and Davide Scaramuzza.
\newblock High speed and high dynamic range video with an event camera.
\newblock \emph{IEEE transactions on pattern analysis and machine intelligence}, 43\penalty0 (6):\penalty0 1964--1980, 2019.

\bibitem[Sagan(2012)]{sagan2012space}
Hans Sagan.
\newblock \emph{Space-filling curves}.
\newblock Springer Science \& Business Media, 2012.

\bibitem[Taguchi et~al.(2010)Taguchi, Agrawal, Ramalingam, and Veeraraghavan]{taguchi2010axial}
Yuichi Taguchi, Amit Agrawal, Srikumar Ramalingam, and Ashok Veeraraghavan.
\newblock Axial light field for curved mirrors: Reflect your perspective, widen your view.
\newblock In \emph{2010 IEEE Computer Society Conference on Computer Vision and Pattern Recognition}, pages 499--506. IEEE, 2010.

\bibitem[Tedaldi et~al.(2016)Tedaldi, Gallego, Mueggler, and Scaramuzza]{tedaldi2016feature}
David Tedaldi, Guillermo Gallego, Elias Mueggler, and Davide Scaramuzza.
\newblock Feature detection and tracking with the dynamic and active-pixel vision sensor (davis).
\newblock In \emph{2016 Second International Conference on Event-based Control, Communication, and Signal Processing (EBCCSP)}, pages 1--7. IEEE, 2016.

\bibitem[Tulyakov et~al.(2021)Tulyakov, Gehrig, Georgoulis, Erbach, Gehrig, Li, and Scaramuzza]{tulyakov2021time}
Stepan Tulyakov, Daniel Gehrig, Stamatios Georgoulis, Julius Erbach, Mathias Gehrig, Yuanyou Li, and Davide Scaramuzza.
\newblock Time lens: Event-based video frame interpolation.
\newblock In \emph{Proceedings of the IEEE/CVF conference on computer vision and pattern recognition}, pages 16155--16164, 2021.

\bibitem[Tulyakov et~al.(2022)Tulyakov, Bochicchio, Gehrig, Georgoulis, Li, and Scaramuzza]{tulyakov2022time}
Stepan Tulyakov, Alfredo Bochicchio, Daniel Gehrig, Stamatios Georgoulis, Yuanyou Li, and Davide Scaramuzza.
\newblock Time lens++: Event-based frame interpolation with parametric non-linear flow and multi-scale fusion.
\newblock In \emph{Proceedings of the IEEE/CVF Conference on Computer Vision and Pattern Recognition}, pages 17755--17764, 2022.

\bibitem[Wan et~al.(2022)Wan, Dai, and Mao]{wan2022learning}
Zhexiong Wan, Yuchao Dai, and Yuxin Mao.
\newblock Learning dense and continuous optical flow from an event camera.
\newblock \emph{IEEE Transactions on Image Processing}, 31:\penalty0 7237--7251, 2022.

\bibitem[Wu et~al.(2017)Wu, Masia, Jarabo, Zhang, Wang, Dai, Chai, and Liu]{wu2017light}
Gaochang Wu, Belen Masia, Adrian Jarabo, Yuchen Zhang, Liangyong Wang, Qionghai Dai, Tianyou Chai, and Yebin Liu.
\newblock Light field image processing: An overview.
\newblock \emph{IEEE Journal of Selected Topics in Signal Processing}, 11\penalty0 (7):\penalty0 926--954, 2017.

\bibitem[Yang et~al.(2002)Yang, Everett, Buehler, and McMillan]{yang2002real}
Jason~C Yang, Matthew Everett, Chris Buehler, and Leonard McMillan.
\newblock A real-time distributed light field camera.
\newblock \emph{Rendering Techniques}, 2002\penalty0 (77-86):\penalty0 2, 2002.

\bibitem[Yu et~al.(2024)Yu, Ren, Han, Wang, Liang, and Shi]{yu2024eventps}
Bohan Yu, Jieji Ren, Jin Han, Feishi Wang, Jinxiu Liang, and Boxin Shi.
\newblock Eventps: Real-time photometric stereo using an event camera.
\newblock In \emph{Proceedings of the IEEE/CVF Conference on Computer Vision and Pattern Recognition}, pages 9602--9611, 2024.

\bibitem[Zhao et~al.(2024)Zhao, Sheng, Yang, Wang, Cong, Cui, Chen, Wang, Wang, Huang, et~al.]{zhao2024survey}
Mingyuan Zhao, Hao Sheng, Da Yang, Sizhe Wang, Ruixuan Cong, Zhenglong Cui, Rongshan Chen, Tun Wang, Shuai Wang, Yang Huang, et~al.
\newblock A survey for light field super-resolution.
\newblock \emph{High-Confidence Computing}, page 100206, 2024.

\bibitem[Zhou et~al.(2021)Zhou, Gallego, Lu, Liu, and Shen]{zhou2021event}
Yi Zhou, Guillermo Gallego, Xiuyuan Lu, Siqi Liu, and Shaojie Shen.
\newblock Event-based motion segmentation with spatio-temporal graph cuts.
\newblock \emph{IEEE transactions on neural networks and learning systems}, 34\penalty0 (8):\penalty0 4868--4880, 2021.

\end{thebibliography}
}

\clearpage
\setcounter{page}{1}
\setcounter{section}{0}
\setcounter{figure}{0}
\renewcommand{\figurename}{Supp. Figure}
% \crefname{figure}{Supp Fig.}{Supp Figs.}
% \Crefname{figure}{Supp Figure}{Supp Figures} 
\maketitlesupplementary

\section{Calibration details for galvanometer setup}

% Event fields result in high-speed, high spatio-angular, HDR light fields. 
% Post-capture refocusing is one of the fundamental applications of light field. 
% We can achieve that using the equations described in \cref{ssec:lf_background}.
% To validate its effectiveness, we conduct tests on two static scenes, each with a foreground object and background elements.
% We test the refocusing ability on static and dynamic scenes (\cref{fig:sec06_04_scan_frequency_test}) using the galvanometer setup. 

The galvanometer scans on a Lissajous curve, and we need to register the exact location of the galvanometer corresponding to each event time stamp to know the light field view we are measuring. One way to achieve this is to synchronize the event clock with the signal sent to the Galvanometer. However, this complicates the hardware design and generates additional events for the synchronization clock, decreasing the overall useful bandwidth. Instead, we took an algorithmic approach to achieve this synchronization using the captured event field data itself. 

\begin{figure}[t]
    \centering
    \includegraphics[width=\linewidth]{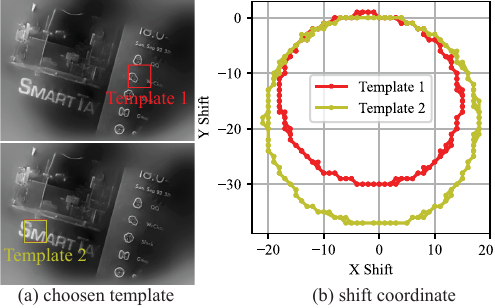}
    \caption{\textbf{Result of template matching.} 
    (a) Two templates are selected from the reconstructed light field image—one in the foreground and one in the background. A template matching algorithm is applied to each template across all light field images to determine the shift amounts in (x,y).
    (b) The calculated shift amounts are plotted, forming a circular pattern consistent with the scanning curve, validating our concept.}
    \label{fig:template_matching}
\end{figure}

The key idea for the algorithmic approach is the observation that any patch in the reconstructed sequence of images will best correlate with other images in the sequence along the scanned Lissajous curve path. For example, we use circular Lissajous curves for scanning in Supp. \cref{fig:template_matching}. For two patches, we show the shifts for best-correlated patches for all subsequent reconstructed frames and notice that the shifts lie on a circle. The size of the circle (or, in general, the size of the Lissajous curve) depends on the depth of the patch. Therefore, by computing the view of the first frame (i.e., its location on the scanning curve), we calibrate the views for the entire light field sequence. This calibration can be accomplished from just one patch and by correlating with as few as three images for circular Lissajous curves, though we used 40 images for robustness. 

\begin{figure}[t]
    \centering
    \includegraphics[width=\linewidth]{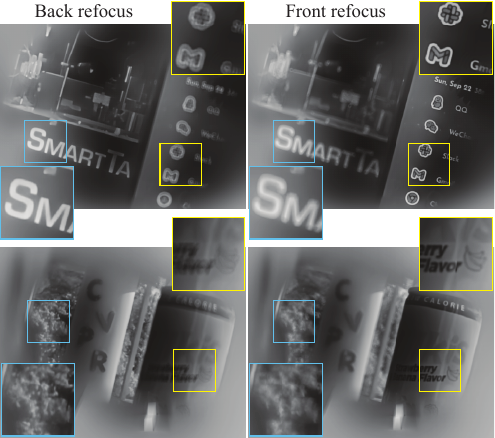}

    \caption{\textbf{Static scene refocusing.} Two distinct static scenes are captured using our galvanometer design at a scan speed of 1 Hz. E2VID is used to reconstruct 100 light field frames. For each scene, we present the reconstructed refocusing result on two selected regions. }
    \label{fig:sec06_01_static_scene_refocus}
\end{figure}

In Supp. \cref{fig:sec06_01_static_scene_refocus}, we use this procedure to calibrate the light field views and use the equations in \cref{ssec:lf_background} to refocus the reconstructed light field at different depths. 
We show two images corresponding to the back refocus and front refocus of the light field data for two static scenes along the rows. In the main manuscript, we also showed refocusing results for dynamic scenes. (see \cref{fig:sec06_02_dynamic_scene_refocus} and \cref{fig:sec06_04_scan_frequency_test})

%%Quinton's text
% To calibrate our galvanometer setup, we perform the refocusing ability on static scene.
% The galvanometer performs a 1~Hz scan along a \textit{circular Lissajous curve} over a 1-second period. 
% We generate 100 light field frames, along the curve, from the acquired event field data using E2VID~\cite{rebecq2019high}.
% In our first attempt, the reconstructed images are treated as an \textit{unstructured light field} \cite{davis2012unstructured}. The template for searching and the shift amount based on the template searching result is shown in Supp. \cref{fig:supp_01}.
% We align the frames by estimating the required shift for a specified patch using a basic pattern-matching algorithm. 
% After computing the cross-correlation, each video frame is shifted accordingly and subsequently summed. This process yields results shown in top row of Supp. \cref{fig:sec06_01_static_scene_refocus}, demonstrating the robustness of our method.

\section{Depth Calibration Details}

Event fields capture multi-view images and hence capture true-depth information about the scene. As mentioned in \cref{subsec:instant-depth}, we use depth from focus to capture the depth of each pixel in the scene. The depth from focus gives us disparity (i.e., the size of the Lissajous curve) for each pixel, and as mentioned in the previous section, the size of the Lissaujous curve is proportional to the patch depth. We compute the proportionality constant by calibrating the system. 

\begin{figure}[t]
    \centering
    \includegraphics[width=\linewidth]{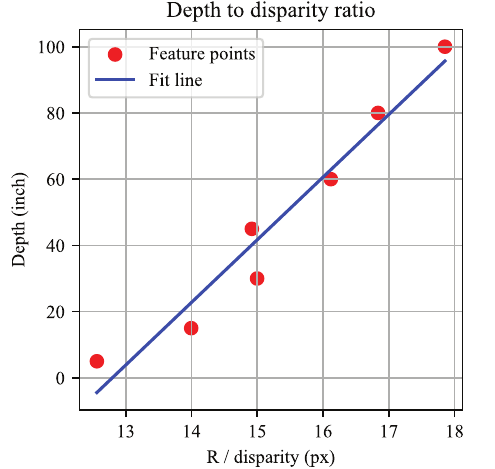}

    \caption{\textbf{Depth calibration.} We calibrate 7 depth points using an LED light placed at distances ranging from 15 inches to 100 inches. The depth (in inches) is linearly fitted to the disparity (in pixels) within this range, serving as our calibrated ratio. For new scenes, we detect disparity using a depth-from-focus algorithm and convert it into real-world depth using this calibrated relationship.}
    \label{fig:depth_calibration}
\end{figure}

We placed a single bright point light source (LED) at a known depth from the galvanometer and computed the size of the Lissaujous curve. We repeated this experiment for different depth locations of the light source and plotted them as shown in Supp. \cref{fig:depth_calibration}. The slope of this line gives us the proportionality constant, and the y-intercept (depth $=$ 0) gives us the bias which is the distance between the camera and the galvo location in pixels.

% We performed depth calibration for our galvanometer design using a similar method. However, instead of a random scene, we captured a very dark environment with a single bright point light source (an LED in this case). By placing the LED bulb at various distances from the galvanometer, we observed different disparities (shift amounts), as illustrated in Supp. \cref{fig:template_matching}. For each depth point, we calculated the approximate shift radius and fitted their relationship into a linear equation, as shown in Supp. \cref{fig:depth_calibration}. 
% We notice that the fitted line intersects at depth $=$ 0 when the shift amount is 13 pixels, indicating that the camera's view is rotating due to the galvanometer's motion.

\section{Codes and Videos}
Along with this supplementary PDF, we provide additional materials to support and reproduce the results presented in the paper, including a short video summarizing the paper's contributions, as well as the code and datasets used to generate the results:  

\begin{itemize}
    \item \textbf{Images and Videos}: All images and videos from the main paper and supplementary materials are included separately in the ``./results/'' folder.
    \item \textbf{Refocusing Algorithms and Raw Data}: The code for the refocusing algorithms, based on our kaleidoscope and galvanometer designs, is available on Anonymous GitHub \cite{eventfield2024}. Additionally, we provide a short version of the raw data corresponding to the galvanometer design showcased in \cref{fig:sec06_02_dynamic_scene_refocus}, as well as the kaleidoscope design highlighted in the slow-motion refocus example in \cref{fig:sec01_01_teaser}. We will make all these public in the non-anonymous version of the manuscript. 
\end{itemize}

% \section{Show the Simulation is Correct}
% {\color{warning} This experiment is not done.}

% \section{Different Lissajous Curve}
% {\color{warning} This experiment is not done.}

\end{document}